\documentclass[11pt,a4paper]{article}
\usepackage[hyperref]{emnlp-ijcnlp-2019}
\usepackage{times}
\usepackage{latexsym}

\usepackage{url}
\usepackage{url}
\usepackage{times}
\usepackage{soul}
\usepackage{url}
\usepackage{amsmath}
\usepackage{graphicx}
\usepackage{algpseudocode}
\usepackage{algorithm}
\usepackage{multirow}

\aclfinalcopy 

\title{Single Training Dimension Selection for Word Embedding with PCA}
\author{Yu Wang \\
  Apple \\
  {\tt w.y@apple.com} \\}

\date{}

\begin{document}
\maketitle
\begin{abstract}
In this paper, we present a fast and reliable method based on PCA to select the number of dimensions for word embeddings. First, we train one embedding with a generous upper bound (e.g. 1,000) of dimensions. Then we transform the embeddings using PCA and incrementally remove the lesser dimensions one at a time while recording the embeddings' performance on language tasks. Lastly, we select the number of dimensions while balancing model size and accuracy. Experiments using various datasets and language tasks demonstrate that we are able to train 10 times fewer sets of embeddings while retaining optimal performance. Researchers interested in training the best-performing embeddings for downstream tasks, such as sentiment analysis, question answering and hypernym extraction, as well as those interested in embedding compression should find the method helpful.
\end{abstract}

\section{Introduction}
Word embeddings constitute an integral part of the implementation of numerous NLP tasks ranging from sentiment classification ~\cite{sentimentWE}, nationality classification ~\cite{nationality}, classification of behavior on Twitter \cite{polarization}, to measuring document similarity ~\cite{documentWE}. The strength of word embeddings stems from their embedding words into low dimensional continuous vector space ~\cite{embeddingWithNoise,glove,line,pca}.

Various algorithms have been proposed for learning words' vector space representations, including most notably Mikolov et al.~\shortcite{word2vec}, Pennington et al.~\shortcite{glove}, and Nickel and Kiela~\shortcite{hierarchicalEmbedding}. However, the exact definition of `low' dimensionality is rarely explored. As pointed out by Yin and Shen~\shortcite{dimensionality}, the most frequently used dimensionality is 300, largely due to the fact that the early influential papers~\cite{word2vec,glove} used 300. Other often used dimensionalities include 200~\cite{line,limitedMemory,hierarchicalEmbedding} and 500~\cite{relationalLearning,deepwalk}.

The impact of vector dimension on embeddings' performance is well known ~\cite{howto}. With too few dimensions, the model will underfit; with too many dimensions the model will overfit. Both undercut the embeddings' performance ~\cite{dimensionality}. What is also known is that the size of the embeddings will grow linearly with the vector dimension ~\cite{limitedMemory,pca}. What is less known is how to identify the optimal vector dimension given any dataset. The method we propose here helps fill this gap.

\begin{figure}[!htbp]
\centering
\includegraphics[width=7.8cm]{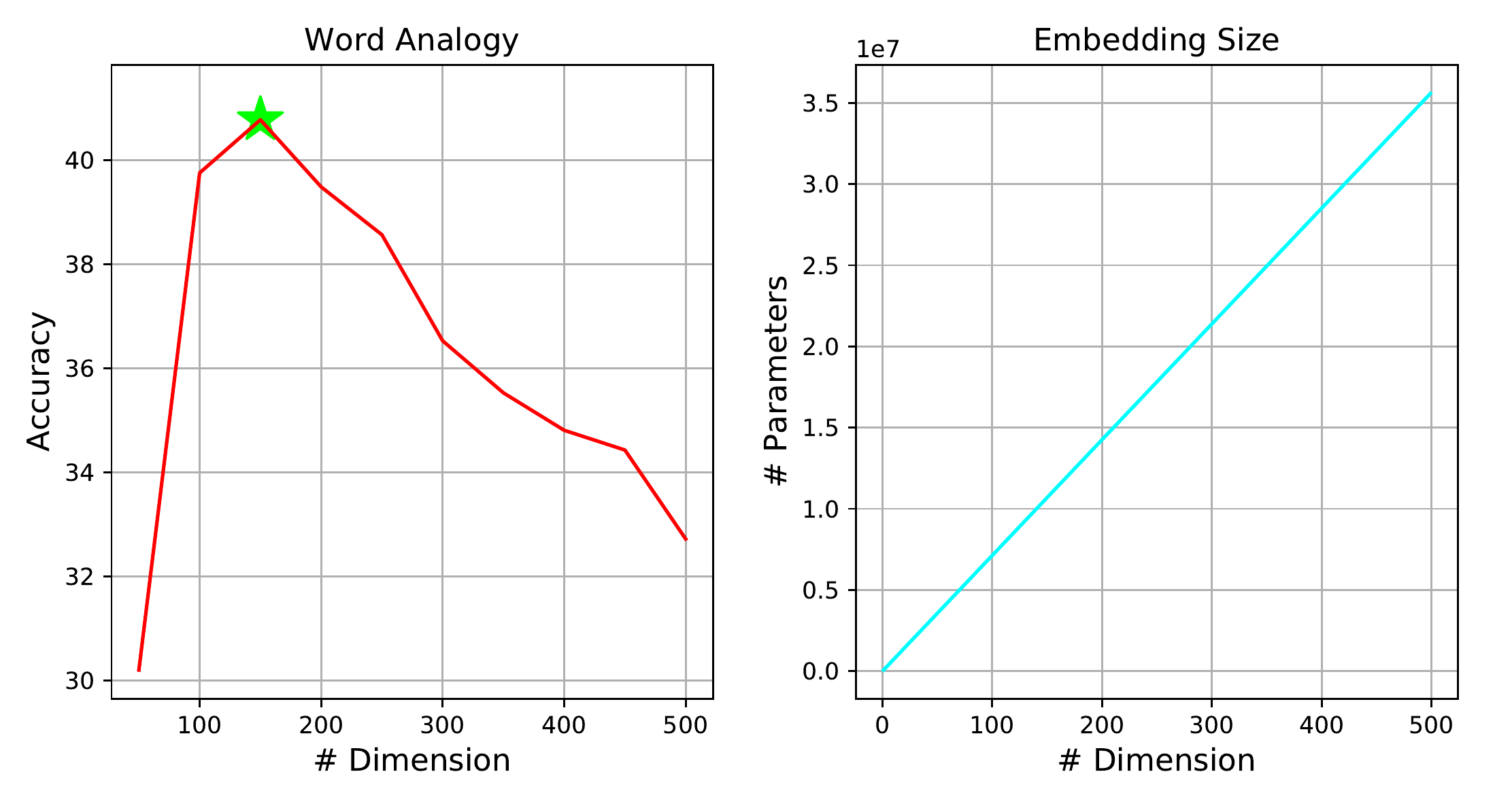}
\caption{Left: Accuracy for the word analogy task as a function of vector dimension (dimension incremental is 50). Right: The number of embedding parameters as a function of vector dimension. Results are generated using the Text8 dataset with a vocabulary size of 71,291.}
\label{demo}
\end{figure}

We offer a fast and reliable PCA-based method that (1) only needs to train the embeddings once and (2) is able to select vector dimension with competitive performance.\footnote{As such, the selected dimension is a Pareto-optimal dimension that balances embedding size and performance.} First, we train one embedding with a generous upper bound (e.g. 1,000) of dimensions. Then we transform the embeddings using PCA and incrementally remove the lesser dimensions one at a time while recording the embeddings' performance on language tasks. Lastly we calculate the best dimensionality and return the corresponding embeddings.

\begin{algorithm*}[!t]
  \caption{Select the Optimal Dimensionality using PCA}
  \begin{algorithmic}
   \State Set dimension upper bound N; select language task from \{word analogy, similarity\}
   \State Train embedding $E$ with N dimensions
   \State Transform $E$ using  PCA: $(u_1, u_2, ..., u_N; \tilde{E}) \leftarrow \mathrm{PCA}(E)$, where $u_1, u_2, .., u_N$ are the new basis \State \:\: \: \:\: \:\:\: vectors, $\tilde{E}$ represents the transformed coefficients
    \For  {$i=N$ to 2} 
    \State  $E$ = $E$-$\tilde{E}_{:,i} \cdot u_i$, where $\tilde{E}_{:,i} $ represents the $i$th column of $\tilde{E}$ and each scalar in $\tilde{E}_{:,i} $ scales vector $u_i$
    \State Evaluate $E$ on the selected language task, record (i, metric)
     \EndFor
  \State Return the selected dimension: [$\underset{i}{\mathrm{arg max}} f$(i, metric)]-1, where $f$ is a score function that balances
\State  \:\: \: \:\: \: \:\: performance and model size and i is between 2 and N
\end{algorithmic}
\end{algorithm*}

Experiments using various datasets and language tasks reveal three key observations:
\begin{itemize}
\item The optimal dimensionality calculated on the  basis of PCA agrees with that by grid search.
\item The resulting embedding is competitive against the one selected by grid search.
\item Different upper-bound dimensionalities (e.g. 500, 1000) point to the same optimal dimensionality.
\end{itemize}
Researchers interested in downstream tasks, such as sentiment analysis ~\cite{sentimentWE}, question answering ~\cite{bert} and hypernym extraction \cite{wrautoencoder}, as well as those interested in embedding compression should find the method helpful.

\section{Related Work}
Our work draws inspiration from Yin and Shen~\shortcite{dimensionality}. The authors build on the Latent Semantics Analysis (LSA) approach and slide $k$ from a lower bound (e.g. 10) to a generous upper bound (e.g. 1,000) in $E=U_{1:k}D^{\alpha}_{1:k,1:k}$, where U and D come from the singular-value decomposition of the signal matrix and $\alpha$ is a hyperparameter to be tuned. For each \textit{k}, the authors generate one corresponding embedding and compare it with the simulated oracle embedding. The \textit{k} that yields the smallest loss is selected. In a similar vein, our work bypasses the problem of training multiple embeddings, often necessitated by grid search, by sliding over all the k values of PCA. Compared with Yin and Shen \shortcite{dimensionality}, our method is easier to implement, as we do not rely on, e.g, Monte Carlo simulations of the oracle embeddings.

At a deeper level, our work is also connected to Yin and Shen~\shortcite{dimensionality} in terms of the trade-off between bias and variance. Yin and Shen~\shortcite{dimensionality} propose pairwise inner product (PIP) loss to measure the quality of an embedding. They decompose the PIP loss into a bias term and a variance term, where reducing the dimension increases the bias term but reduces the variance. They show that the bias-variance trade-off reflects the signal-to-noise ratio in dimension selection. While there is no exact 1-1 mapping from their theorem to our work, we do have analogous discussion in Section 3. The PCA step in our algorithm enables us to identify and drop dimensions that (1) contribute less to the explained variance in the embedding and yet (2) contribute equally to cosine similarity. In essence, our PCA step is removing dimensions with low signal-to-noise ratios.

Our work also draws strength from the literature on post-processing embeddings. 
Mu and Viswanath ~\shortcite{allbuttop} demonstrate that removing the top dominating directions in the PCA-transformed embeddings helps improve the embeddings' performance in word analogy and similarity tasks. Building on that, Raunak~\shortcite{pca} shows that by performing another iteration of PCA and dropping the bottom directions, one can further improve a model's performance as well as reduce its size. Both works focus on improving pre-trained embeddings's performance in terms of accuracy and size. By contrast, our algorithm selects the optimal dimensionality before the actual training.

In addition, our work is related to a few recent studies on model compression ~\cite{sparseEmbedding,limitedMemory,compressingEmbeddings}. In particular, Ling et al.~\shortcite{limitedMemory} seek to drop the least significant digits to reduce the embeddings' size. By comparison, our method removes the least significant dimensions (in terms of explained variance ~\cite{prml}). It should be noted that the two methods complement each other, as one focuses on dimension selection whereas the other on limited precision representation.

\section{Algorithm}

In this section, we formally describe how to select a competitive dimensionality by training one embedding. We state the proposed algorithm in Algorithm 1.

First, we note that the PCA transformation (when retaining all the N dimensions) does not affect embeddings' performance on word similarity tasks. Any potential performance gain should come from dropping the lesser dimensions. By ``lesser dimensions,'' we mean the dimensions that contribute little to the explanation of variance ~\cite{prml}.

\begin{figure}[!h]
\centering
\includegraphics[width=7.5cm]{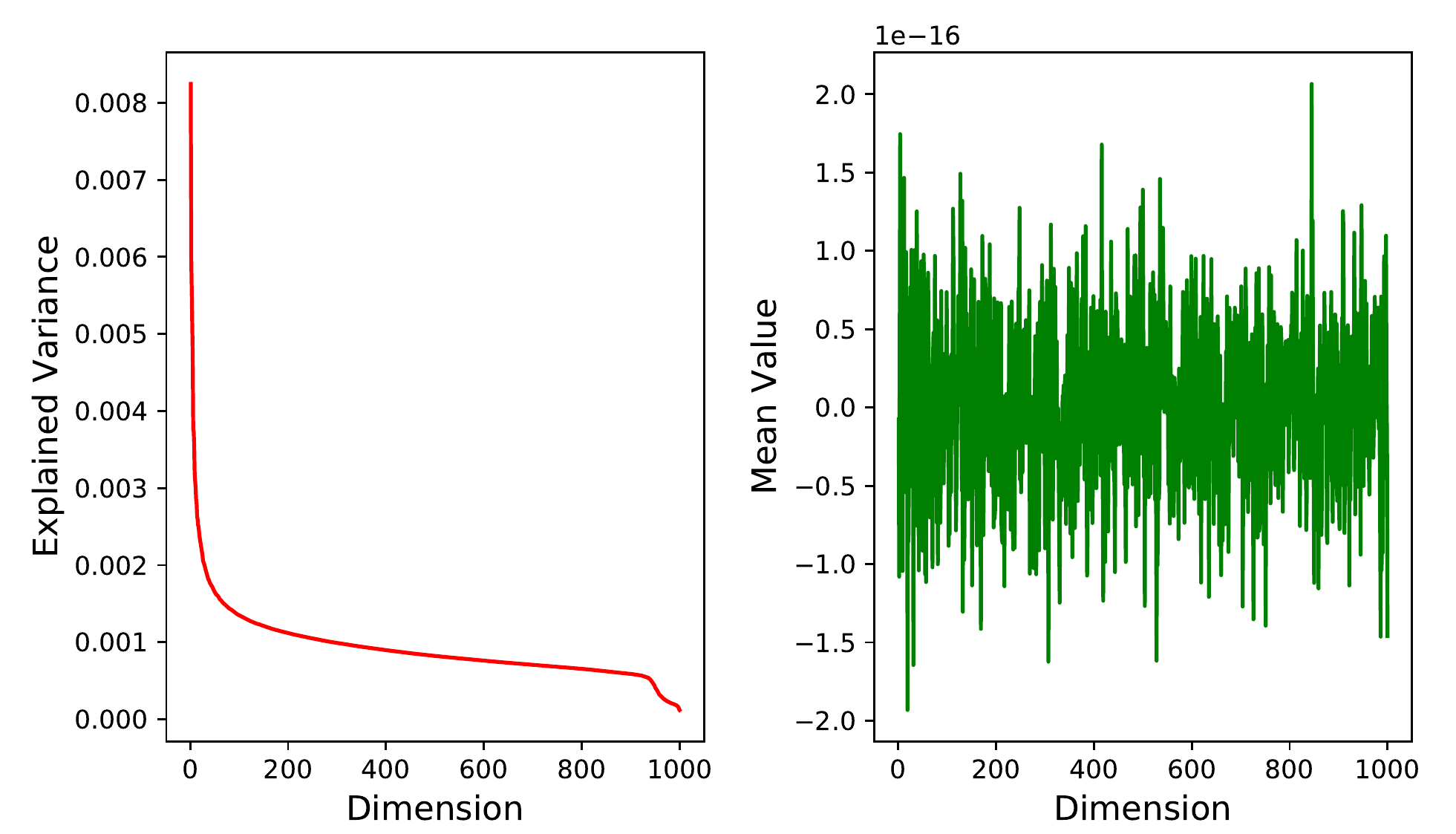}
\caption{Left: explained variance drops sharply after the top 100 dimensions. Right: all the 1,000 dimensions  have roughly the same mean value.}
\label{pca_metric}
\end{figure}

\begin{figure*}[h]
\hspace*{0.3cm} 
\includegraphics[width=15cm]{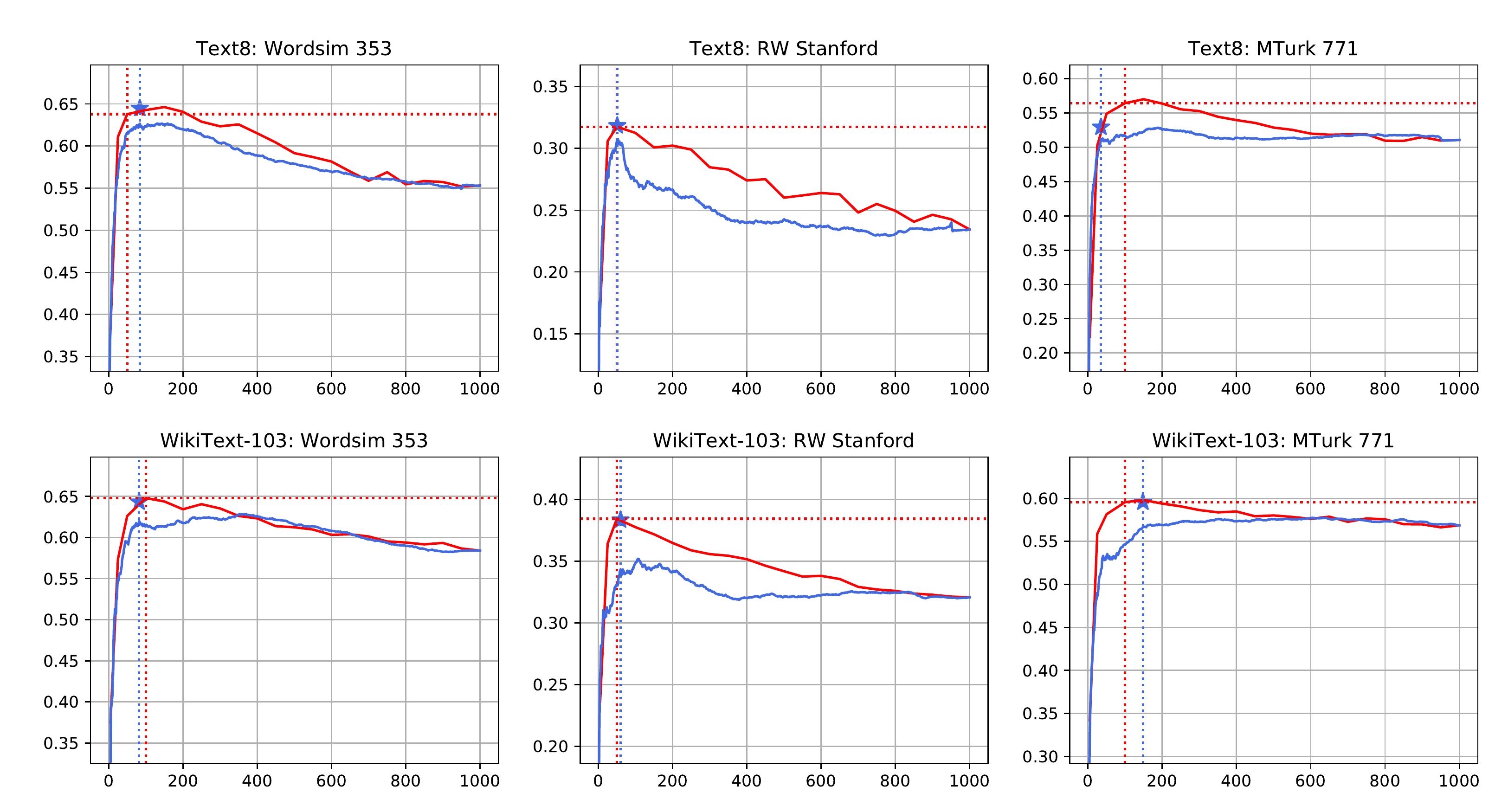}
\caption{Across different benchmarks and different training datasets, the optimal dimensionality that our method (blue curve) identifies closely matches grid search (red curve). The top row is based on Text8. The bottom row is based on WikiText-103. The vertical dotted lines represent the optimal dimensionality for the respective curves. The horizontal dotted line represents the performance of grid search and the blue star marks the performance of our method. The score function is \textit{f}(i, metric)=  metric-50$\times$i. All the curves are averaged over 5 random runs.}
\label{gtruth}
\end{figure*}

In Figure \ref{pca_metric}, we first transform an embedding using PCA, so that each new dimension represents a principal component. We show that the explained variance goes drastically down for dimensions beyond the 100th and stays relatively stable for 100th-1,000th dimensions. In terms of magnitude, the first dimension explains 69.2 times more variance than the last dimension.

While different dimensions contribute differently to the explained variance, they nonetheless contribute equally to the calculation of inner product (Figure \ref{pca_metric}, right). Therefore, the lesser dimensions, with less variance but equal weighting, effectively decreases the discriminative power of the model. Removing these lesser dimensions enables us to focus on the more discriminative dimensions. To identify optimal dimensionality, beyond which all dimensions are considered lesser, we turn to experiments in Section \ref{Experiments}.

\section{Experiments}\label{Experiments}
Given the popularity of the word2vec model ~\cite{word2vec}, we use Skip-gram as the embedding algorithm. Following Yin and Shen ~\shortcite{dimensionality} and Grave et al.~\shortcite{ccache}, we use the widely used benchmark datasets, Text8 ~\cite{text8} and WikiText-103~\cite{wiki103}, as the training datasets. For ground truth, we train 20 embeddings, with dimensions ranging from 50 to 1,000 at an interval of 50 as well as two embeddings with dimensions of 5 and 25.\footnote{A larger interval will save researchers more time, but may result in a sub-optimal dimensionality. While a smaller interval may give more accurate results, it requires training more sets of embeddings proportionally. Our proposed method, by contrast, can scan through all the dimensions at the finest granularity, 1, at virtually no extra cost.} Here we have made the implicit assumption that 1,000 is an upper bound for the embedding space's dimensionality. For each embedding, we train 200 epochs ~\cite{glove,swivel} and keep only the checkpoint that performs best on the word analogy task ~\cite{embedding}. Our experiments focus on (1) comparing our method with grid-search based ground truth and (2) examining consistency between different upper bounds. 

\begin{figure*}[!h] 
\centering
\includegraphics[width=15cm]{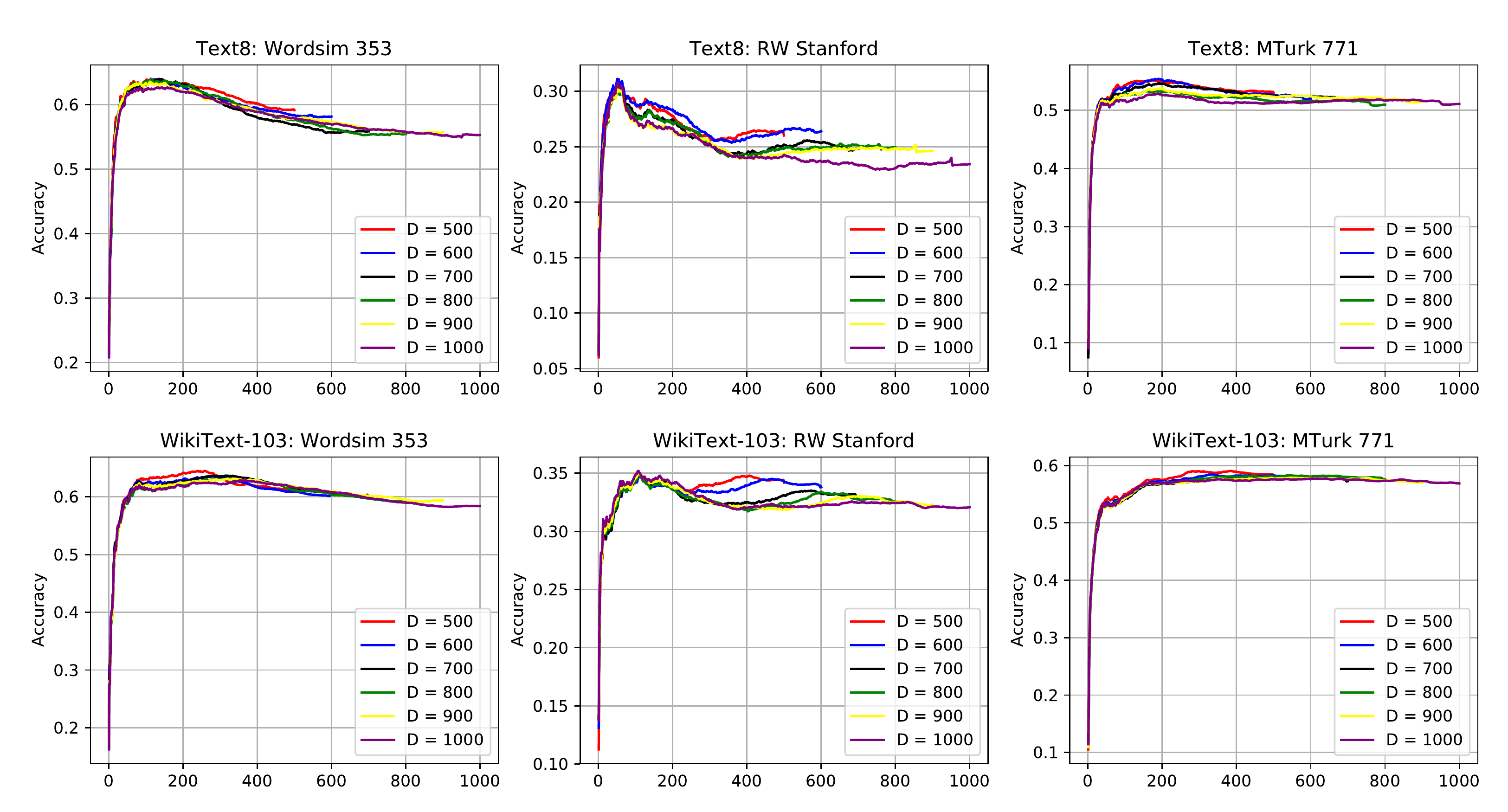}
\caption{Across different benchmarks and different training datasets, consistency is observed when the upper bound is set to 500, 600, 700, 800, 900 and 1,000. The curves trace closely each other, in particular for WordSim 353 and MTurk 771. The top row is based on the Text8 dataset. The bottom row is based on the WikiText-103 dataset. All the curves are averaged over 5 random runs. Best viewed in color.}
\label{robust}
\end{figure*}
\subsection{\hspace{-0.1cm}Performance Compared with Grid Search}
In this subsection, we compare the optimal dimensionality that our method calculates with the ground truth (Figure \ref{gtruth}). We perform the comparison across three testing datasets: Wordsim 353~\cite{wordsim}, RW Stanford~\cite{rw_stanford} and MTurk 771~\cite{mturk771}. Figure \ref{gtruth} demonstrates that our PCA based method (with one training only) is able to uncover the optimal dimensionality. Table \ref{distance} reports the distance in selected dimensionalities.

\begin{table}[]
\caption{Distance in Dimensionalities between Grid Search and Our Proposed Method (Numbers within the bucket size 50 are in bold)}
\label{distance}
\renewcommand{\arraystretch}{1}
\begin{tabular}{|l|l|l|l|}
\hline
Datasets         & WordSim     & RW Stanford  & MTurk       \\ \hline
Text8            & \textbf{34} & \textbf{1}   & 65          \\ \hline
WikiText         & \textbf{18} & \textbf{10}  & \textbf{48} \\ \hline
\textit{Average} & \textbf{26} & \textbf{5.5} & 66.5        \\ \hline
\end{tabular}
\end{table}

\begin{table}[!h]
\small
\caption{Performance (\textit{Correlation}) Comparison between Grid Search and Our Proposed Method}
\label{comparison}
\setlength{\tabcolsep}{1pt}
\renewcommand{\arraystretch}{1.2}
\begin{tabular}{|c|c|l|l|l|}
\hline
Datasets                      & Method      & \multicolumn{1}{c|}{WordSim}  &  \multicolumn{1}{c|}{RW Stanford} &  \multicolumn{1}{c|}{MTurk}  \\ \hline
\multirow{2}{*}{Text8}        & G.S. & 63.8        & 31.8        & 56.4      \\ \cline{2-5} 
                              & Ours  & 64.5 (101.1\%)        & 31.8(100.3\%)        & 52.9 (93.8\%)     \\ \hline
\multirow{2}{*}{WikiText} & G.S. & 64.8        & 38.4        & 59.6      \\ \cline{2-5} 
                              & Ours  & 64.3  (99.3\%)      & 38.3 (99.7\%)       & 59.6 (100\%)      \\ \hline      
                              \textit{Average}                      & -      & 100.2\% &  100\% &  96.9\%  \\ \hline             
\end{tabular}
\end{table}	

We also observe that the optimal embedding that results from our method (with retraining) is competitive against the optimal embedding found using grid search. In Figure \ref{gtruth}, we mark out the respective optimal performances of the two approaches in similarity tasks. In Table \ref{comparison}, we further report the optimal performance achieved by grid search and our method as well as their relative performance. Even though we have only trained one embedding (and one retraining), our method, on average, is able to achieve 100.2\% (WordSim 353) to 96.9\% (MTurk 771) of the optimal performance by grid-searching through 22 sets of embeddings.

\subsection{Consistency across Upper Bounds}
One hyperparameter involved in our method is the upper bound. Intuitively, we expect the upper bound should be higher for larger datasets. In this subsection, we demonstrate our method is robust against different upper bounds. 

In Figure \ref{robust}, we vary the dimension from 500 to 1,000 at an increment of 100. We observe that the dimensionality our method selects is consistent across different upper bounds. Based on the demonstrated consistency, different upper bounds can be selected and still the optimal dimensionality can be uncovered as long as the chosen upper bound is larger than the optimal dimensionality.

\subsection{Efficiency Compared with Grid Search}
\begin{figure}[!htbp]
\centering
\includegraphics[width=7.5cm]{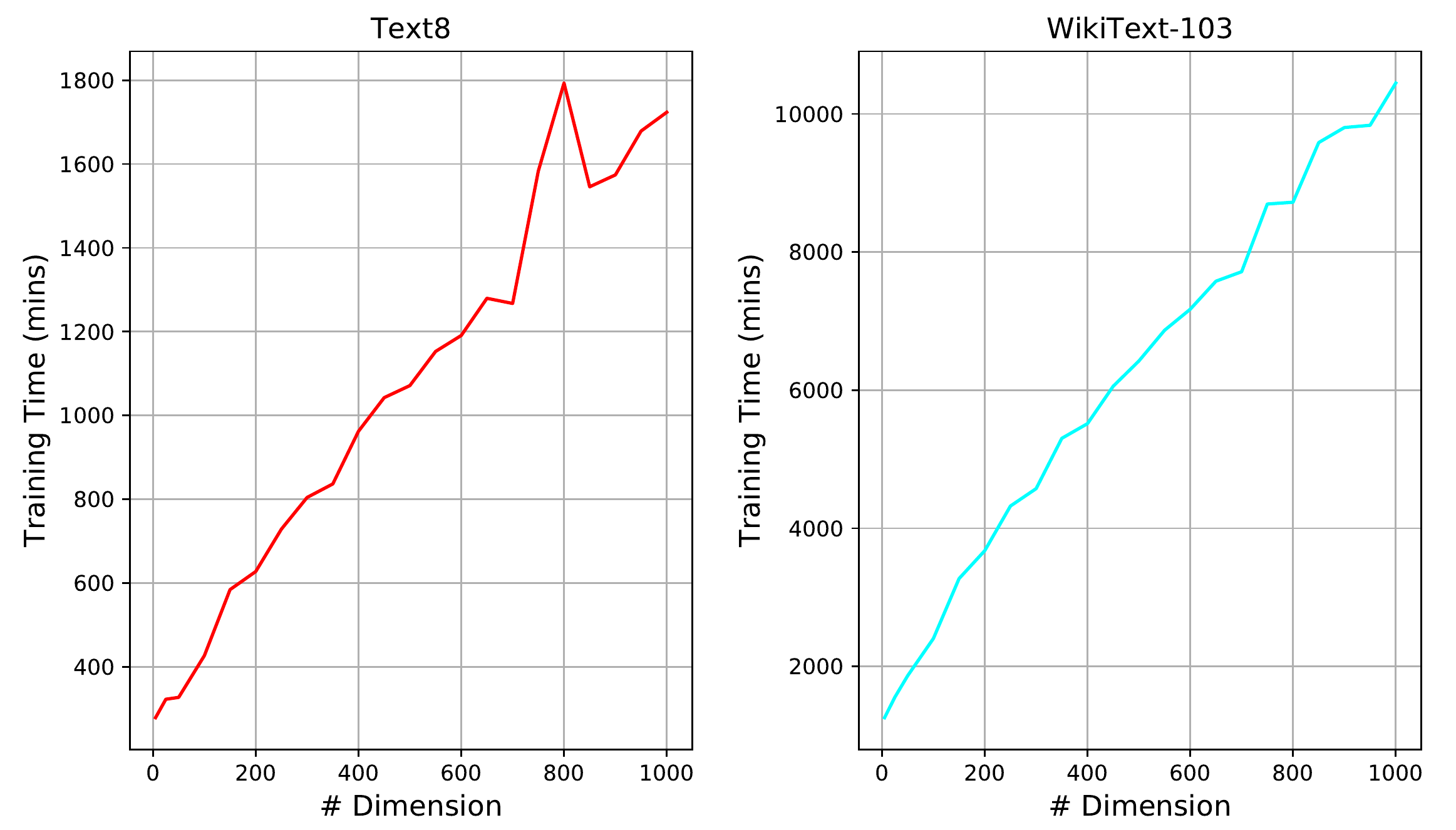}
\caption{Left: training time for Text8 for embeddings of different dimensions for 200 epochs using 6 CPUs. Right: training time for WikiText-103 for embeddings of different dimensions for 200 epochs using 10 CPUs. All results are averaged over 5 random runs.}
\label{timing}
\end{figure}

In this subsection, we  report the running time of our algorithm and compare it with that of grid search. We have recorded the running time of our experiments and that of the PCA transformations. We average them over 5 random runs (Figure \ref{timing}).

For Text8, grid search takes 22,801 minutes cumulatively. Training a 1,000-dimension embedding takes 1,724 minutes. The PCA step takes 22 minutes (note that for each embedding we only need one PCA operation). This represents a 13.1x speedup for our method. For WikiText-103, grid search takes 132,652 minutes. Training a 1,000-dimension embedding takes 10,448 minutes. The PCA step takes 34 minutes. This represents a 12.7x speedup.

We note that the comparison results are dependent on grid granularity. A coarser grid search could save researchers more time, at the cost of performance loss. 

\section{Conclusion}
In this paper, we provided a fast and reliable method based on PCA to select the number of dimensions for training word embeddings. First, we train one embedding with a generous upper bound (e.g. 1,000) of dimensions. Then we transform the embedding using PCA and incrementally remove the lesser dimensions while recording the embeddings' performance on language tasks.  Experiments demonstrate that (1) our method is able to identify the optimal dimensionality, (2) the resulting embedding has competitive performance against grid search, and (3) our method is robust against the selection of the upper bound.

\section*{Acknowledgement}
The authors would like to thank Hang Zhao and Srinivasan Venkatachary for supporting this project, would like to thank Russ Webb, Arnab Ghoshal, and Jaewook Chung for their insightful comments on early versions of this paper, and would like to thank the anonymous EMNLP-IJCNLP reviewers for their reviews and suggestions.

\bibliography{./../citation/citations}
\bibliographystyle{acl_natbib}
\end{document}